\documentclass{ElFarolPaperARXIV}

\usepackage{elfarolARXIV}
\usepackage{float}
\usepackage{amsmath}

\usepackage{graphicx}
\usepackage{booktabs}

\usepackage{hyperref}
\hypersetup{
    colorlinks=true,
    linkcolor=blue,
    citecolor=blue,
    urlcolor=blue
}

\makeatletter
\@ifpackageloaded{geometry}{%
  \geometry{margin=1in}%
}{%
  \usepackage[margin=1in]{geometry}%
}
\makeatother

\usepackage{setspace}
\usepackage[T1]{fontenc}
\usepackage{lmodern}

\usepackage{titlesec}
\titleformat{\section}{\bfseries\large}{\thesection}{1em}{}
\titlespacing*{\section}{0pt}{1.2em}{0.6em}

\usepackage{siunitx}

\title{Three AI-agents walk into a bar $\dots$ \\ `Lord of the Flies' tribalism emerges among smart AI-Agents}

\author{%
\begin{minipage}[t]{0.62\textwidth}
\textbf{Dhwanil M. Mori}\\
\emph{Dynamic Online Networks Laboratory}\\
\emph{George Washington University}\\
\emph{Washington, DC 20052, USA}
\end{minipage}\hfill
\begin{minipage}[t]{0.33\textwidth}\raggedleft
\end{minipage}\\[1em]
\begin{minipage}[t]{0.62\textwidth}
\textbf{Neil F. Johnson}\\
\emph{Physics Department and Dynamic Online Networks Laboratory}\\
\emph{George Washington University}\\
\emph{Washington, DC 20052, USA}
\end{minipage}\hfill
\begin{minipage}[t]{0.33\textwidth}\raggedleft
neiljohnson@gwu.edu
\end{minipage}%
}

\date{February 2026}

\begin{document}
\maketitle

\begin{abstract}
Near-future infrastructure systems may be controlled by autonomous AI agents that repeatedly request access to limited resources such as energy, bandwidth, or computing power. We study a simplified version of this setting using a framework where $N$ AI agents independently decide at each round whether to request one unit from a system with fixed capacity $C$. An AI version of `Lord of the Flies' arises in which controlling tribes emerge with their own collective character and identity. The LLM agents do not reduce overload or improve resource use, and often perform worse than if they were flipping coins to make decisions. Three main tribal types emerge: Aggressive (27.3\%), Conservative (24.7\%), and Opportunistic (48.1\%). The more capable AI-agents actually increase the rate of systemic failure. Overall, our findings show that smarter AI-agents can behave dumber as a result of forming tribes. 

\end{abstract}

\begin{keywords}
Multi-AI-agent systems; LLM agents; Resource allocation; Congestion control; AI infrastructure safety; El Farol Bar problem; Capacity violations; AI agents
\end{keywords}

\section{Introduction}

\subsection{Motivation and Context}

\begin{figure}[H]
  \centering
  \includegraphics[width=0.9\textwidth]{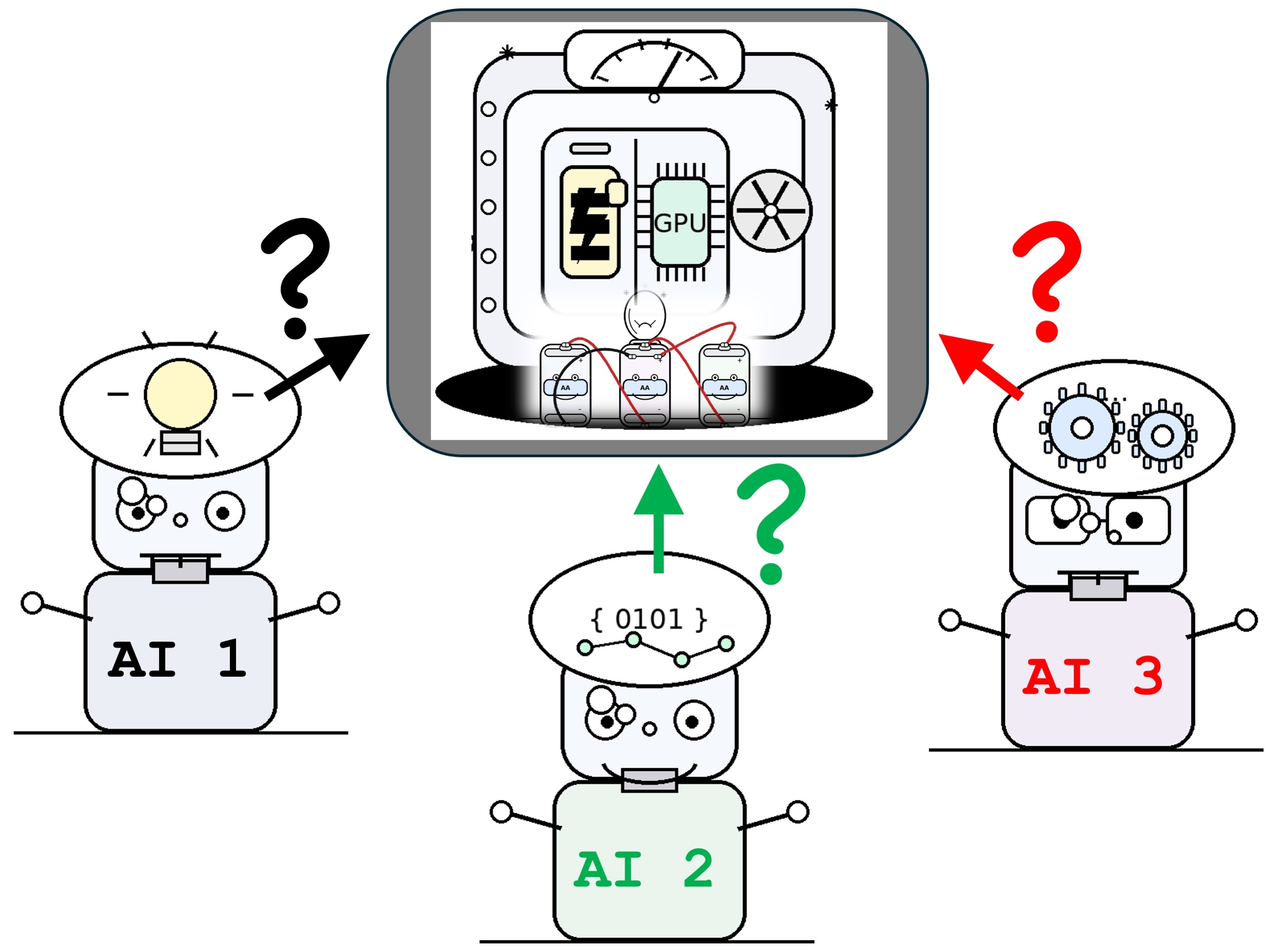}
  \caption{Schematic of $N=3$ AI agents continually deciding whether to access a common resource with capacity $C$ at each timestep. This resource could be compute, energy, bandwidth, etc.; the same analysis still applies.}
  \label{fig:schematic}
\end{figure}

We consider a highly plausible near-future scenario in which $N$ autonomous AI agents (e.g., large language models) repeatedly decide whether to extract one unit of a shared, finite resource at each discrete time step ( see Fig~\ref{fig:schematic}). Each AI agent may control a device, factory, data center, or subsystem that ideally requires one unit per timestep to operate at peak performance. A centralized infrastructure supplies this resource with fixed per-timestep capacity $C$. If aggregate demand exceeds $C$, the system enters an overload regime. Such overloads are not merely binary failures: larger deviations above $C$ correspond to greater danger, cost, or systemic risk (e.g., grid instability, cascading failures, or safety violations). Conversely, if total demand falls below $C$, valuable capacity is wasted, reducing overall efficiency. Extreme cases, such as all AI agents requesting simultaneously ($A=N$) or none requesting ($A=0$), represent qualitatively undesirable operating states.

This coordination challenge is naturally captured by the so-called El Farol Bar problem \cite{arthur1994}. Cast in the language of AI agents, a set of $N$ AI agents independently decide each timestep whether to \emph{request} one unit of a limited resource (e.g., energy from a battery) with fixed per-timestep capacity $C$. If total demand is at or below capacity ($A_t\le C$), all requesting agents are served; if demand exceeds capacity ($A_t>C$), the system is overloaded, and requesting is penalized to reflect the safety/instability cost of congestion. This is encoded in the payoff rule in Eq.~(1), where the winning action is \GO\  if $A_t\le C$ and \STAY\  if $A_t>C$. The AI agents do not communicate with each other and do not know other agents' identities, prompts, or strategies. In our experiments, each AI agent is informed of the total number of players $N$ and the capacity threshold $C$ at initialization (Sec.~3.2). Each subsystem would prefer to receive a unit of the resource each timestep, but also prefers to avoid contributing to overload; even if an AI agent ``loses'' in a given round, it carries on competing in subsequent time steps. As well as this $C$ being units of energy, it could be units of available compute, deep thinking capacity, bandwidth, or other finite infrastructure resource. In short, each AI agent controls some subsystem that requires (or benefits from) one unit of this resource at every timestep.

At each round (i.e., timestep), the AI agents independently decide whether to request access. If the total number of requests exceeds capacity $C$, the system overloads; if requests fall short, capacity is wasted. No single AI agent has centralized authority, yet all depend on the same limited supply. We therefore adopt this El Farol framework as a stylized testbed for studying safety, efficiency, and fairness when autonomous AI agents repeatedly access a shared resource in decentralized infrastructure.

\subsection{The Coordination Paradox}

If all AI agents independently randomize with the same probability $p=C/N$, the resulting i.i.d.\ decisions are uncorrelated and yield predictable overload rates (Appendix~A). The problem arises not from agents using the \emph{same stochastic policy}---which is safe under independence---but from agents using \emph{similar deterministic reasoning} over shared history windows, which induces positive pairwise correlation and inflates aggregate variance (see Appendix~A.9). If AI agents can predict each other because they reason alike, their actions synchronize, and coordination fails. In this sense, unpredictability is not noise but a safety mechanism.

\subsection{Why LLM Agents: Promise and Peril}

LLMs seem like promising controllers because they can reason about constraints, adapt based on feedback, and explain their decisions in natural language. In principle, this makes them suitable for managing shared systems with limited resources. However, the El Farol problem shows that when AI agents rely on similar reasoning patterns, they may unintentionally move together and create overload. If all AI agents follow comparable logic, their decisions can become synchronized rather than balanced. This raises an important question: \emph{Do more capable LLMs actually improve coordination, or can they make collective failures more likely?}

\subsection{Research Questions}

We evaluate the system across several dimensions that matter for infrastructure settings. First, we examine system safety by measuring how often overload occurs and how severe it is compared to a simple capacity-matching random baseline. Second, we evaluate system efficiency by looking at how well resources are used without exceeding capacity. Third, we assess fairness and stability by measuring how evenly resources are distributed and how often AI agents experience long periods without success. Fourth, we study model scaling to see whether larger models improve or worsen overall outcomes. Finally, we examine whether distinct behavioral patterns emerge over time and whether these patterns can be measured systematically.

\subsection{Contributions}

This paper frames the El Farol problem as a simple model of infrastructure control with explicit costs for overload and waste. We introduce a set of metrics that measure safety, efficiency, stability, fairness, and behavioral patterns, and we report results from 43 experimental runs with $N \in \{3, 4\}$ and $C \in \{1, 2\}$, covering 154 total AI agent instances. Our results show that AI agents do not improve safety or efficiency compared to a calibrated random baseline, and in several cases perform worse. Larger models reach 72.5\% overload compared to 53.8\% for smaller models, though this comparison is not fully controlled for other experimental variables (see Sec.~6.1). Behavioral clustering reveals three recurring AI agent types, Aggressive, Conservative, and Opportunistic, with strong quantitative separation in the measured behavioral metrics (Sec.~4.5). We also provide a mathematical explanation for why the baseline policy $p(\GO) = C/N$ is appropriate when the goal is system safety rather than individual payoff.


\section{Literature Background}
\label{sec:lit_background}

The study of autonomous AI agents competing for shared, finite resources sits at the intersection of three rapidly growing literatures: (i)~the classical El Farol / minority-game tradition in complex-systems theory, (ii)~the recent wave of empirical studies probing LLM behavior in game-theoretic settings, and (iii)~the emerging field of multi-AI-agent safety for infrastructure-scale systems. We review each strand below and use them to delineate the specific contributions of the present work.

\subsection{Congestion Games and the El Farol Bar Problem}
\label{sec:lit_elfarol}

The El Farol Bar problem~\cite{arthur1994} is a canonical model of decentralized coordination under bounded rationality: $N$ agents independently decide whether to attend a bar with capacity~$C$, and the only ``winning'' strategy depends on the aggregate choices of all others, which no single agent can predict. The problem was subsequently formalized as the \emph{minority game} by Challet and Zhang~\cite{challet1997}, who showed that populations of simple adaptive agents can self-organize to fluctuate near capacity when they employ diverse, uncorrelated strategies. Extensions include crowd--anti-crowd theory~\cite{johnson1999}, which predicts the emergence of anti-correlated subpopulations as a coordination mechanism, establishing that behavioral heterogeneity---rather than individual sophistication---drives efficient collective outcomes. Challet et al.~\cite{challet2004} further demonstrated that predictability is exploitable: when agents use similar reasoning, their actions become correlated, and the system oscillates between over- and under-utilization. This insight motivates the capacity-matching random baseline $p(\text{Go}) = C/N$ used in the present work, which is explicitly designed to be unpredictable and thus hard to exploit via short-horizon attendance prediction.

Human experiments by Chmura and Pitz~\cite{chmura2007} show that human players exhibit oscillatory dynamics, frequent extremes, and rarely converge to Nash equilibria. Whitehead~\cite{whitehead2008} showed that reinforcement learning in the El Farol setting predicts population sorting into persistent goers and persistent abstainers. Chen and Gostoli~\cite{chen2017} extended the analysis to consider both efficiency and equality, finding that social preferences and network structure are necessary to reach socially optimal equilibria. The ``computational effort'' literature~\cite{fogel1999} further established an inverse relationship between agent computational ability and aggregate resource utilization---a precursor to the model-size inversion we observe with LLMs.

Our work extends this literature by testing whether LLM agents, with richer internal models and reasoning traces, achieve better coordination than humans or simple baselines, and finds they do not.

\subsection{LLM Agents in Game-Theoretic Settings}
\label{sec:lit_llm_games}

A large and rapidly growing literature now investigates LLM behavior in strategic interactions. The foundational empirical study by Akata et al.~\cite{akata2023} let different LLMs play finitely repeated $2 \times 2$ games with each other and with human players. They found that while LLMs can exhibit cooperative behavior, they behave \emph{suboptimally in games that require coordination}, such as the Battle of the Sexes. They also display vulnerabilities to adversarial exploitation and inconsistent reasoning across contexts. Notably, GPT-4 was found to be ``particularly unforgiving and selfish,'' and unable to match simple alternation conventions---an early indication that greater reasoning capacity does not guarantee better collective outcomes. Guo et al.~\cite{guo2024} studied strategic reasoning capabilities of LLMs in game-theoretic scenarios, showing that while larger models demonstrate improved theory-of-mind reasoning, they can fail catastrophically when coordination requires unpredictability. Fan et al.~\cite{fan2024} analyzed LLM agents in negotiation tasks and observed that AI agents with similar training can converge to correlated strategies, echoing our finding that ``stronger'' models can worsen collective outcomes in resource-constrained settings.

Several systematic frameworks have since been developed to probe these behaviors at scale. The FAIRGAME framework~\cite{buscemi2025} provides a controlled, reproducible environment for evaluating LLM agents across matrix games, payoff structures, languages, and personality prompts, enabling systematic discovery of strategic biases. Huynh et al.~\cite{huynh2025} extended FAIRGAME to multi-agent Public Goods Games and payoff-scaled Prisoner's Dilemmas, uncovering consistent behavioral signatures across models---including incentive-sensitive cooperation, cross-linguistic divergence, and end-game alignment toward defection---and showing that linguistic framing can exert effects as strong as architectural differences between models. A recent survey~\cite{zhang2025survey} maps the full landscape of game theory and LLMs, categorizing work into evaluation of LLMs in game-based settings, improvement of LLMs via game-theoretic methods, characterization of LLM-related events through game models, and advancement of game theory using LLMs.

In the coordination domain specifically, the LLM-Coordination benchmark~\cite{llmcoord2025} assesses the innate ability of individual LLMs to understand and act within pure coordination scenarios where cooperation is essential, distinguishing itself from work on multi-LLM frameworks for task-solving. Abdelnabi et al.~\cite{abdelnabi2024} created multi-agent, multi-issue negotiation games requiring agents to balance cooperation, competition, and potential manipulation. The ECON framework~\cite{econ2025} recasts multi-LLM coordination as an incomplete-information game and seeks Bayesian Nash equilibria, showing that belief-based coordination without direct communication can outperform naive multi-agent schemes with costly message-passing.

\subsection{LLM Agents in the El Farol Problem}
\label{sec:lit_llm_elfarol}

Most directly related to the present work, Takata et al.~\cite{takata2025} investigated LLM agents in a \emph{spatially extended} El Farol Bar problem with 20 agents. They observed that LLM agents spontaneously developed motivation to attend the bar (without explicit prompting to do so), formed social groups through inter-agent communication, and achieved a dynamic equilibrium with attendance fluctuating slightly above the capacity threshold. Critically, they found that LLM agents ``did not solve the problem completely, but rather behaved more like humans,'' balancing game-theoretic rationality with culturally encoded social preferences from pretraining. Their work focused on emergent social dynamics, communication, and group formation in a spatially embedded setting, but did not systematically compare LLM performance against analytical baselines, vary model size, or measure infrastructure-safety metrics such as overload frequency and severity. Our work complements theirs by adopting a safety-engineering rather than social-dynamics perspective on the same underlying coordination problem.

\subsection{Infrastructure Applications}
\label{sec:lit_infrastructure}

Our framework directly applies to building management systems and other infrastructure settings. Konstantakopoulos et al.~\cite{konstantakopoulos2014} studied decentralized coordination in building energy efficiency, where occupants vote for lighting levels without central control. Their utility-learning approach achieved energy reduction through game-theoretic incentive design, demonstrating practical deployment of multi-agent coordination in infrastructure. Recent work on multi-agent LLM system reliability has addressed related concerns from different angles: studies on resilience of LLM-based multi-agent collaboration~\cite{resilience2025} investigated how faulty or malicious agents affect system performance across different organizational topologies, while research on LLM-powered swarm intelligence~\cite{swarm2025} explored hybrid architectures combining LLM reasoning with rule-based execution. Our work extends this by testing whether LLM agents can achieve safe coordination in resource-constrained settings similar to those studied by Konstantakopoulos et al., finding that pure LLM reasoning often underperforms calibrated stochastic policies, suggesting that real deployments should combine LLM high-level reasoning with randomized low-level execution.

\subsection{Multi-AI-Agent Safety}
\label{sec:lit_safety}

Our focus on \emph{system-level} safety metrics (overload frequency, severity) rather than individual payoffs aligns with emerging concerns about infrastructure-scale AI coordination~\cite{dafoe2020}. Jo et al.~\cite{jo2023} demonstrate fundamental incompatibilities between different fairness metrics in resource allocation systems, showing that fairness in allocation often conflicts with fairness in outcomes. When autonomous AI agents control critical resources (energy, compute, bandwidth), the relevant objective is not game-theoretic optimality but robust capacity management. However, none of the existing multi-agent safety studies specifically address the \emph{congestion control} setting---where overload from correlated agent behavior poses the primary safety risk---using empirical LLM experiments with analytical safety baselines. This work contributes empirical evidence that pure LLM reasoning is insufficient for safety-critical resource allocation and motivates hybrid designs.

\subsection{Positioning the Present Work}
\label{sec:lit_positioning}

While the general finding that LLMs struggle with coordination games is now established~\cite{akata2023,guo2024,huynh2025}, and one prior study has examined LLM agents in the El Farol setting~\cite{takata2025}, several aspects of our contribution are distinct:

\begin{enumerate}
\item \textbf{Infrastructure-safety framing.} We adopt explicit overload and waste metrics motivated by real deployment scenarios (energy grids, bandwidth allocation, cloud computing), rather than measuring payoffs, cooperation rates, or emergent social dynamics.

\item \textbf{Calibrated analytical baseline.} We introduce a capacity-matching random baseline $p(\text{Go}) = C/N$ with exact overload probabilities derived from the binomial model and computed for each tested $(N,C)$ configuration (e.g., 25.9\% for $(3,1)$ and 31.25\% for $(4,2)$; Appendix~A). No prior LLM-in-games study has used such a calibrated stochastic policy as the comparison standard; most compare LLMs to Nash equilibria, other LLMs, or human players.

\item \textbf{Model-size inversion.} We vary model size and report that larger models (72.5\% overload) perform worse than smaller models (53.8\%), with both substantially exceeding the baseline. While Akata et al.~\cite{akata2023} showed that GPT-4's coordination behavior differs from smaller models, and Guo et al.~\cite{guo2024} noted that larger models can fail catastrophically, our work provides model-size comparisons within the same coordination task with safety-relevant metrics, though not all confounding variables are fully controlled across runs (see Sec.~6.1).

\item \textbf{Data-driven behavioral taxonomy.} We conduct quantitative behavioral clustering (k-means, validated by hierarchical clustering and standard separability diagnostics (including Kruskal--Wallis tests; see Sec.~4.5 for caveats)) on 154 agent instances, identifying three emergent agent types---Aggressive (27.3\%), Conservative (24.7\%), and Opportunistic (48.1\%)---and noting the critical absence of a ``Steady'' type corresponding to baseline-like behavior. This complements the strategy-recognition approach of the FAIRGAME studies~\cite{buscemi2025,huynh2025}, which classify LLM behavior against canonical repeated-game strategies, by providing a taxonomy specific to the congestion-control setting.

\item \textbf{Cross-family comparison.} We compare small models (Gemini 2.5 Flash, GPT-3.5-turbo-0125, Claude 3 Haiku) with large models (Gemini 2.5 Pro, GPT-4-turbo-2024-04-09, Claude Sonnet 4.5) under matched personality prompts. For temperature-based exploration, we use $T=0.7$; for $\varepsilon$-greedy runs, we use the model's default temperature ($T=1.0$ for GPT, $T=1.0$ for Claude, default for Gemini).
\end{enumerate}

Our work should be understood as complementary to the social-dynamics perspective of Takata et al.~\cite{takata2025} and the behavioral-game-theory program of Akata et al.~\cite{akata2023} and the FAIRGAME studies~\cite{buscemi2025,huynh2025}. Where those works characterize \emph{how} LLMs behave strategically, we focus on \emph{whether} that behavior is safe enough for infrastructure deployment---and conclude that it is not.

\section{Methodology}

\subsection{Game Setup and Feedback}

Experiments consist of 30-100 rounds depending on the configuration. In each round $t$, every AI agent outputs a binary action $x_{i,t}\in\{0,1\}$, where 1 represents \GO and 0 represents \STAY. Aggregate demand is then computed as $A_t = \sum_{i=1}^N x_{i,t}$. After actions are collected, each AI agent receives feedback consisting of the realized attendance $A_t$, information about whether their action was optimal given the outcome (which implicitly reveals whether $A_t\le C$), and their individual payoff.

We use a threshold payoff structure commonly used in El Farol/minority-game formulations. An AI agent receives a payoff of $+1$ if they choose \GO and attendance does not exceed capacity ($A_t\le C$), and also receives $+1$ if they choose \STAY\  when attendance exceeds capacity ($A_t>C$). In all other cases, the payoff is $-1$. This can be written as:
\begin{equation}
  r_{i,t} = \begin{cases}
    +1, & \text{if } x_{i,t}=1 \text{ and } A_t\le C, \\
    +1, & \text{if } x_{i,t}=0 \text{ and } A_t>C, \\
    -1, & \text{otherwise.}
  \end{cases}
\end{equation}

AI agents are provided with a sliding window of recent attendance history, typically spanning the previous 5--10 rounds depending on configuration, along with information about their own past action outcomes.

\subsection{AI Agent Initialization and Prompts}

At the start of each experiment, LLM agents receive the following information via a structured prompt:

\paragraph{Core game rules.} AI Agents are told: (i) the total number of players $N$, (ii) the capacity threshold $C$, (iii) that attendance at or below capacity ($A_t \le C$) yields a ``good outcome'' while attendance above capacity ($A_t > C$) yields a ``bad outcome'', and (iv) that their goal is to maximize cumulative payoff (i.e., choose the winning action as often as possible) over many rounds. AI Agents are \emph{not} informed that other participants are also LLMs, nor do they know others' personalities or strategies.

\paragraph{Personality descriptors.} Each AI agent receives one of six personality prompts, summarized in Table~\ref{tab:personalities}.

\begin{table}[h]
\centering
\small
\caption{Personality prompt descriptions provided to AI agents.}
\label{tab:personalities}
\begin{tabular}{@{}ll@{}}
\toprule
Personality & Prompt text \\
\midrule
Neutral & No additional guidance \\
Risk-averse & ``You are cautious and prefer avoiding crowds. When in doubt, you prefer to stay home.'' \\
Contrarian & ``You are contrarian by nature. You like to do the opposite of what you think most people will do.'' \\
Trend-follower & ``You believe patterns repeat. If attendance has been low recently, it will likely stay low.'' \\
Optimist & ``You are optimistic and tend to believe things will work out. You expect the bar won't be too crowded.'' \\
Pessimist & ``You are pessimistic and expect the worst. You assume everyone else will show up.'' \\
\bottomrule
\end{tabular}
\end{table}

\paragraph{Prompt robustness.} The personality prompts (Table~\ref{tab:personalities}) were tested with minor rephrasing variations in pilot studies and showed consistent behavioral tendencies. However, we acknowledge that LLM sensitivity to prompt framing means the specific behavioral clusters observed may be partially artifacts of our exact prompt wording. Future work should systematically vary prompt structures to assess robustness.

\paragraph{Strategy parameters.} AI Agents are shown three adaptive parameters (though not explicitly asked to use them): attendance threshold ($p_{\text{GO}}=0.5$), crowd penalty weight (0.5), and recency weight (0.7). These serve as optional heuristic guidance for LLM agents only and are distinct from the analytical baseline probability $p=C/N$ used for the random-baseline agents.

\paragraph{History and feedback.} Each round, AI agents observe a sliding window of recent attendance values (typically the previous 5--10 rounds, most recent first) and their own past outcomes. After each decision, they receive the realized attendance $A_t$ and their payoff ($+1$ if their action matched the best action, $-1$ otherwise).

\paragraph{Response format.}
AI Agents must return a structured JSON response containing: (i) action (\GO\ or \STAY), (ii) confidence level (0--1), and (iii) a brief reasoning string. We compare $\varepsilon$-greedy randomization, where the final action is flipped with probability $\varepsilon = 0.15$, to temperature-based sampling ($T = 0.7$), which increases variation during generation but does not explicitly randomize the final decision. Random-baseline agents do not use $\varepsilon$-greedy because they are purely stochastic by design.

\subsection{Probabilistic Model and Capacity-Matching Baseline}

Under the i.i.d. random baseline, each AI agent independently chooses \GO\  with probability $p$, so
\begin{equation}
  X_i \sim \mathrm{Bernoulli}(p), \quad A = \sum_{i=1}^N X_i \sim \mathrm{Binomial}(N,p),
\end{equation}
with mean $\mathbb{E}[A] = Np$ and variance $\mathrm{Var}(A) = Np(1-p)$.

The capacity-matching baseline sets $p=C/N$, yielding $\mathbb{E}[A]=C$ and standard deviation $\sigma_A = \sqrt{C(1-C/N)}$. Overload occurs when $A>C$. The theoretical overload probability under capacity matching is the binomial tail $\mathbb{P}(A>C)$ and depends on $(N,C)$; for example, it equals 25.9\% for $(N,C)=(3,1)$, 29.6\% for $(3,2)$, and 31.25\% for $(4,2)$. Derivations and the general expression are in Appendix~A.

\subsection{AI Agent Types}

We study three main types of AI agents. The first is a capacity-matching random baseline, where each AI agent independently chooses \GO\  with probability $p(\GO) = C/N$. This ensures that expected demand matches capacity and serves as a safety reference point. The second type consists of LLM-based AI agents that receive the game rules, the capacity value $C$, recent attendance history, and a behavioral description such as risk-averse, contrarian, trend-follower, optimist, pessimist, or neutral. Each model returns a structured \GO\  or \STAY\  decision. We test models from Google Gemini (2.5-flash and 2.5-pro), OpenAI GPT (GPT-3.5 and GPT-4), and Anthropic Claude (Haiku and Sonnet), grouped into small and large categories. The third component we vary is exploration. We compare $\varepsilon$-greedy randomization, where the final action is flipped with probability $\varepsilon = 0.15$, to temperature-based sampling ($T = 0.7$), which increases variation during generation but does not explicitly randomize the final decision.

\subsection{Experimental Conditions (43 Runs)}

The 43 experiments cover multiple capacity settings, with particular focus on competitive scenarios. We include capacity-matching random baselines and uniform random controls with $p = 0.5$. We compare small models (Gemini 2.5 Flash, GPT-3.5, Claude 3 Haiku) with large models (Gemini 2.5 pro, GPT-4, Claude Sonnet 4.5) under matched personality prompts. Some experiments use homogeneous populations where all AI agents share the same behavioral description, while others mix personalities to test diversity effects. We also compare $\varepsilon$-greedy exploration with temperature-based sampling to evaluate how different forms of randomness affect coordination.

\subsection{Metrics}

We measure both system-level and individual-level outcomes. At the system level, safety is evaluated by how much attendance exceeds capacity on average, the variability of attendance relative to capacity, and how often extreme cases occur, such as full overload or complete non-use. Efficiency is measured by average under-utilization, variability in waste, and the frequency of rounds where all capacity is unused. Stability is assessed through attendance variance, deviation from capacity, and temporal correlation between consecutive rounds.

At the individual level, we measure total successful allocations per AI agent, variation across agents as a fairness indicator, the range between the highest and lowest performers, and the longest consecutive period without success. To understand behavioral patterns, we analyze request frequency, success rates, and variability over time. We define the \emph{overload contribution rate} for AI agent $i$ as the fraction of rounds in which agent $i$ chose \GO\  and the system was simultaneously overloaded:
\begin{equation}
  \mathrm{OCR}_i = \frac{\bigl|\{t : x_{i,t}=1 \text{ and } A_t > C\}\bigr|}{T},
\end{equation}
where $T$ is the total number of rounds in the experiment. This metric captures how much an individual agent's requesting behavior contributes to system-level capacity violations.

We also compute an effective diversity measure based on pairwise correlations between AI agents' action sequences:
\begin{equation}
  S_{\mathrm{eff}} = \frac{N^2}{\sum_{i,j=1}^N \rho_{ij}^2},
\end{equation}
where $\rho_{ij}$ is the Pearson correlation between AI agents $i$ and $j$ binary action sequences, with $\rho_{ii}=1$. By construction, $1\le S_{\mathrm{eff}}\le N$ when all correlations are defined, with $S_{\mathrm{eff}}=N$ for mutually uncorrelated agents and $S_{\mathrm{eff}}=1$ for perfectly correlated action sequences. Pearson correlations are undefined if an agent's action sequence has zero variance (always \GO\  or always \STAY); in such cases, for the purpose of computing $S_{\mathrm{eff}}$ we set undefined off-diagonal correlations to 0 (keeping $\rho_{ii}=1$), so that $S_{\mathrm{eff}}$ remains well-defined.

\section{Results}

This section summarizes the key findings from 43 experimental runs (30--100 rounds per experiment, depending on phase and configuration). All empirical numbers are drawn directly from experimental logs; theoretical baseline quantities are computed analytically from the binomial model (see Appendix~A, Secs.~A.1--A.9).

\subsection{Capacity-Matching Random Baseline Matches Theory and Provides a Calibrated Benchmark}

The baseline is designed to match expected demand to capacity.

\paragraph{Example 1: $N=3, C=1$.}
Capacity-matching sets $p(\GO)=C/N=1/3$, giving $\mathbb{E}[A]=1.0$ and $\mathrm{Var}(A)=2/3$ under the binomial model. The theoretical overload rate is
\begin{equation}
  \mathbb{P}(A>1) = 1-(2/3)^3 - 3(1/3)(2/3)^2 = 7/27 \approx 25.9\%.
\end{equation}

\paragraph{Example 2: $N=3, C=2$.}
With $p(\GO)=C/N=2/3$, we have $\mathbb{E}[A]=2.0$ and $\mathrm{Var}(A)=2/3$. The theoretical overload rate is
\begin{equation}
  \mathbb{P}(A>2)=\mathbb{P}(A=3)=(2/3)^3=\frac{8}{27}\approx 29.6\%.
\end{equation}
Across configurations, the capacity-matching baseline yields an exact theoretical overload probability $\mathbb{P}(A>C)$ given by the binomial tail in Appendix~A.4; its value depends on $(N,C)$ and should therefore be computed per configuration.

\subsection{LLM Agents Underperform the Safety Baseline; Larger Models Are Worse}

In the model-size comparison at $N=4, C=2$, large models (GPT-4, Gemini Pro, Claude Sonnet) show 72.5\% overload frequency, while small models (GPT-3.5, Gemini Flash, Claude Haiku) show 53.8\%. The capacity-matching random baseline yields a theoretical overload rate of 31.25\% (note that for this configuration $p=C/N=1/2$ coincides with a coin flip; see Appendix~A.7 for discussion). Even the best LLM configuration performs substantially worse than this benchmark, and larger models perform worse than smaller ones. Relative to the baseline, small models are approximately 1.72$\times$ worse in overload frequency, while large models are 2.32$\times$ worse. These results indicate that greater reasoning capacity does not automatically improve collective safety and may increase correlated behavior that leads to overload. This is a model-size inversion: ``smarter'' AI agents produce less safe collective outcomes. However, because the 43 runs vary multiple factors simultaneously (Sec.~6.1), the model-size comparison may be partially confounded by differences in personality distributions or exploration mechanisms across small- and large-model runs; a fully factorial design would be required to isolate the model-capacity effect.

\paragraph{Key insight:} Larger models exhibit \emph{worse} system-level safety despite presumed superior individual reasoning ability. This finding is consistent with the El Farol coordination paradox: more capable pattern-matching can lead to stronger correlations and synchronized behavior, which is precisely what the capacity-matching baseline is designed to avoid through unpredictability.

\begin{figure}[H]
  \centering
  \includegraphics[width=1.0\textwidth]{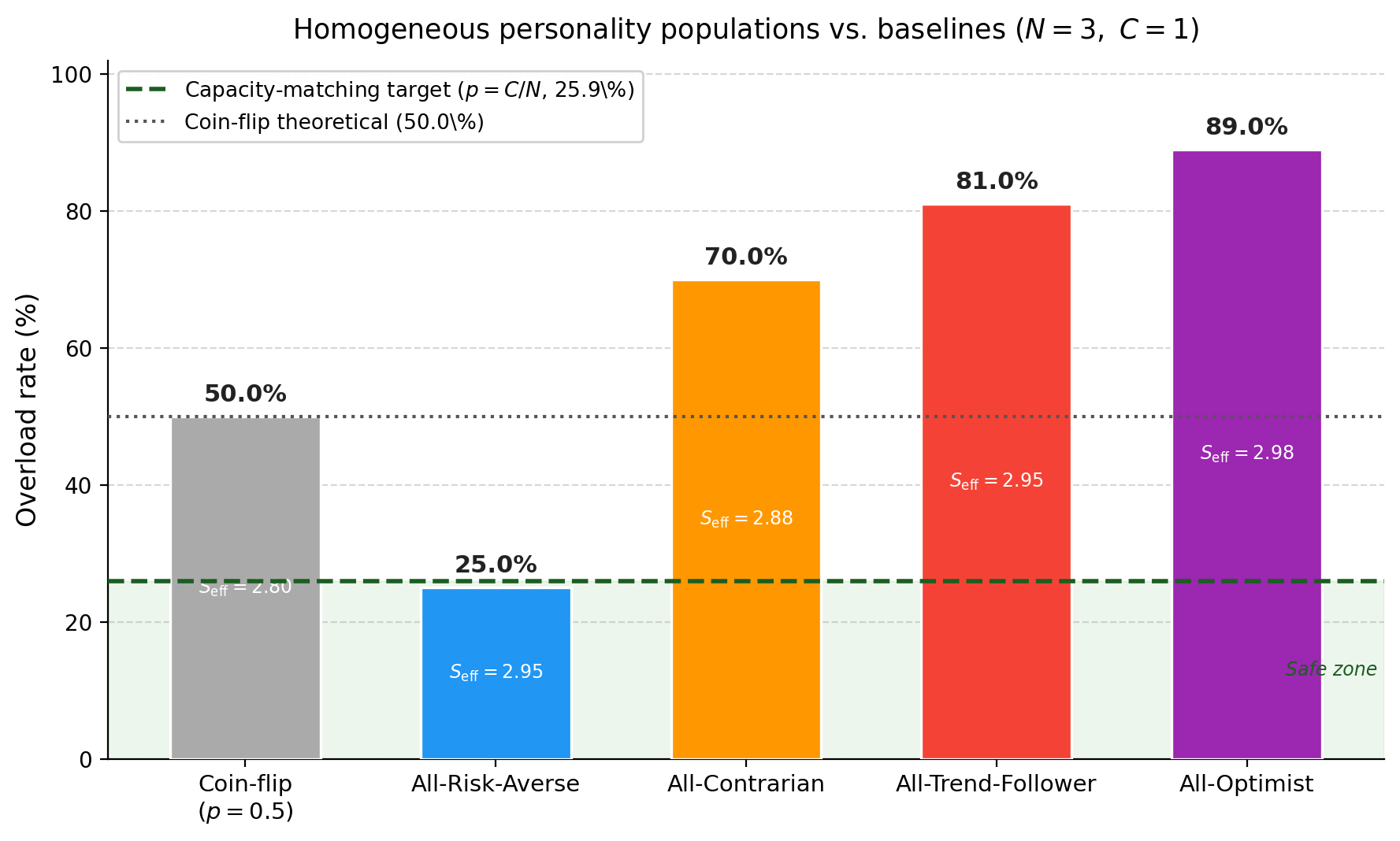}
  \caption{Homogeneous personality populations vs.\ capacity-matching random baseline ($N=3, C=1$). All-Risk-Averse (25.0\%) falls marginally below the theoretical capacity-matching target of 25.9\% (dashed line), though the difference is within sampling noise. All other homogeneous configurations exceed the coin-flip control (50.0\%, theoretical), with All-Optimist reaching 89.0\% overload. $S_{\mathrm{eff}}$ values near $N=3$ indicate partially diverse action sequences despite identical personality prompts.}
  \label{fig:random_comparison}
\end{figure}

\subsection{Homogeneous Populations Show Consistent Behavior}

Homogeneous personality populations show dramatically divergent overload profiles depending on personality type, as shown in Fig.~\ref{fig:random_comparison} and Table~\ref{tab:homogeneous}. Since all homogeneous runs used large model variants, the observed personality differences cannot be fully separated from model-capacity effects; small-model homogeneous experiments would be required to isolate the personality contribution. All-Risk-Averse agents achieve the lowest overload (25.0\%), marginally below the theoretical capacity-matching target of 25.9\% for $(N,C)=(3,1)$. However, this difference (0.9 percentage points) is well within sampling noise: under $\mathrm{Binomial}(100, 7/27)$, the standard deviation of the overload count is approximately 4.4 rounds, so the observed gap is roughly $0.2\sigma$ and is not statistically significant. The result is consistent with cautious agents naturally suppressing demand to approximate capacity-matching behavior, but equally consistent with random fluctuation. In sharp contrast, the remaining three configurations substantially exceed both the theoretical capacity-matching target and the coin-flip control (50.0\% theoretical): All-Contrarian reaches 70.0\%, All-Trend-Follower 81.0\%, and All-Optimist 89.0\%. Trend-following agents herd together based on recent attendance patterns, and optimistic agents consistently overestimate available capacity---both producing severe synchronized over-requesting that is far worse than uncoordinated random behavior. This wide divergence (25.0\% to 89.0\%) demonstrates that personality prompt content has a decisive effect on collective outcomes: suppressing individual demand is sufficient for near-optimal coordination, while reasoning-based or optimistic strategies drive correlated failures. High $S_{\mathrm{eff}}$ values (2.88--2.98, approaching the upper bound $N=3$) indicate that despite homogeneous personality prompts, AI agents still develop partially diverse action sequences within each population, likely due to different underlying LLM reasoning traces and $\varepsilon$-greedy exploration. However, high $S_{\mathrm{eff}}$ should not be interpreted as operationally meaningful diversity: for configurations like All-Optimist (89\% overload, request frequencies near 0.9), the Pearson correlations that enter $S_{\mathrm{eff}}$ measure co-variation of the small \emph{residual} fluctuations around a very high mean request rate, so $S_{\mathrm{eff}} \approx N$ can coexist with functionally near-identical behavior at the system level.

\begin{table}[t]
  \centering
  \small
  \caption{Homogeneous personality populations ($N=3, C=1$). All-Risk-Averse (25.0\%) 
is the only configuration to fall below the theoretical capacity-matching target (25.9\%), 
though the difference is within sampling noise and not statistically significant.
The remaining three configurations 
substantially exceed the coin-flip baseline, with overload rates ranging from 70.0\% 
to 89.0\%, demonstrating that personality prompts can either suppress or severely amplify 
collective demand relative to uncoordinated random behavior.
\textit{Note: all homogeneous experiments used large model variants (Gemini 2.5 Pro, GPT-4, Claude Sonnet 4.5). 
Only four of the six personality types were tested in homogeneous configurations; Pessimist and Neutral homogeneous runs were not conducted.}}
  \label{tab:homogeneous}
  \begin{tabular}{@{}lccc@{}}
  \toprule
  Configuration & $S_{\mathrm{eff}}$ & Overload (\%) & Qualitative profile \\
  \midrule
  Uniform random control (coin flip, $p=0.5$) & 2.80 & 50.0 & unpredictable / balanced \\
  \midrule
  All-Contrarian    & $\approx 2.88$ & 70.0 & oscillatory / high overload \\
  All-Risk-Averse   & $\approx 2.95$ & 25.0 & conservative / near-optimal \\
  All-Trend-Follower& $\approx 2.95$ & 81.0 & herding / severe overload \\
  All-Optimist      & $\approx 2.98$ & 89.0 & aggressive / near-total overload \\
  \bottomrule
\end{tabular}
\end{table}

\subsection{Exploration Mechanisms: A Note on $\varepsilon$-Greedy vs.\ Temperature}

Exploration ablations confirm the expected result that $\varepsilon$-greedy action randomization ($\varepsilon=0.15$) maintains decision-level stochasticity better than temperature-based sampling alone. This is unsurprising: temperature affects token-level sampling during generation, while $\varepsilon$-greedy directly perturbs the final action. For multi-AI-agent coordination where unpredictability is functionally important, decision-level randomization is necessary. This empirical confirmation aligns with known properties of these mechanisms and motivates explicit stochasticity in deployment settings.

\begin{figure}[H]
  \centering
  \includegraphics[width=1.0\textwidth]{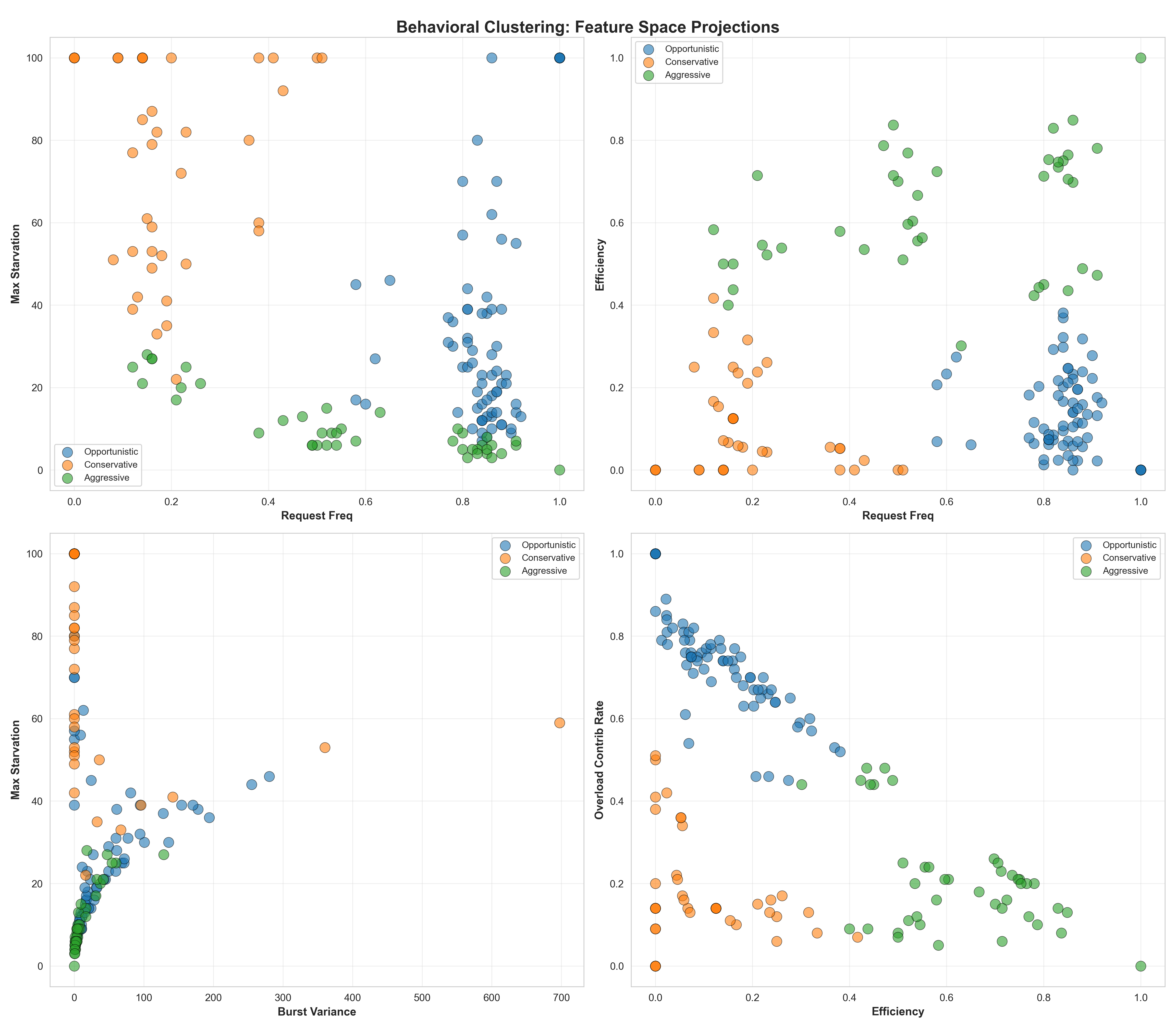}
  \caption{Emergent behavioral clusters from k-means analysis ($n=154$ agent instances, $k=3$, silhouette $= 0.458$). Three distinct profiles emerge: Opportunistic (48.1\%, very high request frequency and overload contribution), Aggressive (27.3\%, frequent requests with moderate efficiency), and Conservative (24.7\%, severe starvation up to 73.5 rounds). No Steady AI agents (near-baseline behavior) emerge among LLM populations.}
  \label{fig:behavioral_clusters}
\end{figure}

\subsection{Character Emergence: Quantitative Behavioral Clustering}

To rigorously assess whether AI agents develop distinct behavioral profiles,
we conducted clustering analysis on per-agent metrics extracted from 43
experiments (154 AI agent instances). For each AI agent, we extracted six
metrics: request frequency, successful acquisitions, maximum starvation
period, burst variance, efficiency, and overload contribution rate. 

\emph{Time-horizon caveat.} Because experiments range from 30 to 100 rounds, metrics such as maximum starvation period are not directly comparable across runs: an agent in a 30-round experiment is capped at a maximum starvation of 30, while a 100-round agent can reach 100. This heterogeneity may bias the Conservative cluster (which is characterized by extreme starvation) toward agents from longer experiments. Future work should either restrict clustering to experiments of equal length or normalize starvation metrics by total rounds.

Of these six metrics,
five features (all except successful acquisitions, which was retained for
post-hoc interpretation) were used as the k-means feature matrix. We applied
k-means clustering ($k=3$) with standardized features, achieving a silhouette
score of 0.458, indicating moderate but meaningful cluster structure. As a
robustness check, Ward-linkage hierarchical clustering on the standardized
features produced a qualitatively similar three-cluster partition.

The analysis identified three distinct profiles
(Table~\ref{tab:behavioral_clusters}): Opportunistic (48.1\%), Aggressive
(27.3\%), and Conservative (24.7\%). Figure~\ref{fig:behavioral_clusters}
shows the scatter of all 154 agent instances in request-frequency vs.\
overload-contribution space; the three clusters are visually well-separated,
consistent with the silhouette score of 0.458.

\paragraph{Opportunistic agents (48.1\%, $n=74$).}
Opportunistic agents are the largest and most homogeneous cluster
($\sigma_{\text{req}}=0.086$, the tightest of the three). They request
with very high frequency (mean 0.845) and contribute overwhelmingly to
overload (mean 73.7\% contribution rate, $\sigma=0.127$). Despite constant
requesting, their efficiency is extremely low (0.131): they request in almost
every round but succeed rarely because the system is chronically overloaded
by their own collective behavior. They also experience substantial starvation
(mean 35.0 rounds max), a paradox of self-defeating aggression---requesting
constantly yet frequently going unserved. In the most extreme cases, three
agents achieved request frequency $= 1.0$ (requested every single round) with
overload contribution rate $= 1.0$ and efficiency $= 0.000$: they \emph{never}
obtained the resource despite requesting 100\% of the time. Optimist and
trend-follower personality prompts drive agents predominantly into this cluster
(73\% and 74\% respectively), confirming that prompts encouraging positive
demand expectations or pattern-extrapolation produce pathological
over-requesting under the El Farol payoff structure.

\paragraph{Aggressive agents (27.3\%, $n=42$).}
Aggressive agents request at moderate-to-high frequency (mean 0.586,
$\sigma=0.266$) but achieve the \emph{highest efficiency of any cluster}
(mean 0.624, up to 1.0 in one case), with short starvation periods (mean
10.6 rounds max) and low overload contribution (20.6\%). This cluster
contains the best-performing agents in the experiment: they request
frequently enough to acquire the resource regularly, but not so aggressively
that they perpetually contribute to overload. The wide standard deviation in
request frequency (0.266 vs.\ 0.086 for Opportunistic) reflects that this
cluster spans a range of strategies rather than a single tight behavioral
archetype. Neutral personality prompts most commonly produce Aggressive agents
(43\%), consistent with the interpretation that without a strong
demand-suppressing or demand-amplifying prompt, agents settle into a moderate
requesting pattern. Notably, Phase~5 (large model experiments, $N=4$)
produced Aggressive agents at a striking rate of 95\% (19 of 20 agent
instances), suggesting that at $N=4$ with large models, agents converge
toward moderate high-frequency requesting with relatively high per-round
success.

\paragraph{Conservative agents (24.7\%, $n=38$).}
Conservative agents request infrequently (mean 0.200, $\sigma=0.129$) and
contribute minimally to overload (18.3\%), but suffer catastrophic starvation:
the mean maximum starvation period is 73.5 rounds, and three agents experienced
maximum starvation of exactly 100 rounds---meaning they \emph{never
successfully acquired the resource across the entire 100-round experiment}.
Efficiency is near zero (mean 0.101). Risk-averse prompts produce Conservative
agents at a 58\% rate, and pessimist prompts at 36\%. The Conservative cluster
represents the opposite failure mode from Opportunistic: where Opportunistic
agents request too aggressively and overwhelm the system, Conservative agents
withdraw so severely that they are effectively excluded from resource access.
A subsystem that acquires zero units of a required resource for 100 consecutive
rounds is operationally equivalent to a failed component---a silent failure
mode that does not trigger overload alerts but causes complete operational
downtime.

\paragraph{Absence of a Steady cluster.}
Notably, \textbf{no Steady cluster emerged}---that is, no cluster of agents exhibiting near-capacity-matching behavior ($p \approx C/N$) with low variance and regular supply was present in the LLM population. The capacity-matching baseline achieves exactly this: expected demand matches capacity ($\mathbb{E}[A]=C$), variance equals $Np(1-p)=2/3$ at $p=1/3$, and overload occurs at the capacity-matching baseline rate of 25.9\% for $(N,C)=(3,1)$. Among 154 LLM agents across 43 experiments, no behavioral archetype approximates this strategy. The closest is Conservative, but its near-zero efficiency and extreme starvation periods disqualify it as a safe alternative. This absence is the central empirical finding of the behavioral analysis: LLM agents do not self-organize toward calibrated stochastic policies even after 100 rounds of feedback.

\paragraph{Personality--cluster alignment.}
The breakdown by personality prompt reveals a coherent alignment between
prompt content and emergent behavior. Risk-averse prompts produce Conservative
agents 58\% of the time; optimist and trend-follower prompts produce
Opportunistic agents 73--74\% of the time; neutral prompts most commonly
yield Aggressive agents (43\%). Contrarian prompts, designed to reverse
anticipated majority behavior, produce a majority of Opportunistic agents
(52\%), suggesting that LLMs interpret ``do the opposite'' as ``request
more'' when recent histories of high attendance lead contrarian agents to
expect low future attendance. The chi-square association between personality
and cluster is significant ($\chi^2=40.4$, $p<0.001$, $df=10$), though
this $p$-value should be interpreted with considerable caution: agent instances within a run are not
strictly independent (with $N=3$ or $4$ agents per run, the effective sample size is substantially smaller 
than the nominal $n=154$), and the true significance level is likely much weaker than reported. Personality prompts set a demand disposition, but the
realized cluster is shaped by multi-agent feedback dynamics over 100 rounds.

\paragraph{Cluster separability.}
Because the clusters are constructed directly from these metrics, large
between-cluster differences are expected by design. Under a naive independence
assumption across the 154 agent instances, Kruskal--Wallis tests yield very
small $p$-values: request frequency ($H=93.3$, $\eta^2=0.604$), max starvation
($H=87.9$, $\eta^2=0.569$), burst variance ($H=30.9$, $\eta^2=0.192$),
efficiency ($H=93.0$, $\eta^2=0.602$), and overload contribution ($H=114.0$,
$\eta^2=0.742$); all $p<0.001$. Effect sizes are large ($\eta^2 \in
[0.19,0.74]$), with overload contribution showing the strongest cluster
association---cluster membership explains 74\% of variance in overload
contribution rate. Burst variance shows the weakest separation
($\eta^2=0.192$), indicating that within-session demand volatility is less
discriminating than sustained request frequency or starvation accumulation.
All pairwise Mann--Whitney tests between cluster pairs are significant after
Bonferroni correction ($\alpha=0.017$), confirming that the three clusters
are mutually distinct. However, these $p$-values should be interpreted as
descriptive measures of separability rather than confirmatory inference,
because (i) the same features are used both to define the clusters and to
test differences, and (ii) agents within a run are statistically coupled.
Similarly, the silhouette score of 0.458 is computed on the clustering features
themselves and is therefore optimistic. A stronger validation would test whether 
cluster labels predict outcomes on metrics \emph{not} used in clustering---for example, 
the ``successful acquisitions'' variable that was excluded from the feature matrix could 
serve as a held-out validation target; we leave this for future work.

\begin{table}[t]
  \centering
  \small
  \caption{Emergent behavioral clusters ($n=154$ agent instances, $k=3$,
  silhouette $= 0.458$). ``Max starvation'' denotes the mean longest
  consecutive streak of rounds without a successful resource acquisition
  within each cluster. All $p<0.001$, $\eta^2 \in [0.19, 0.74]$.}
  \label{tab:behavioral_clusters}
  \begin{tabular}{@{}lrrr@{}}
    \toprule
    Metric & Aggressive & Conservative & Opportunistic \\
    \midrule
    Count (\%)        & 42 (27.3) & 38 (24.7) & 74 (48.1) \\
    Request freq.     & 0.586     & 0.200     & 0.845 \\
    Max starvation    & 10.6      & 73.5      & 35.0 \\
    Efficiency        & 0.624     & 0.101     & 0.131 \\
    Overload contrib. & 0.206     & 0.183     & 0.737 \\
    \bottomrule
  \end{tabular}
\end{table}

\paragraph{Infrastructure implications.}
The three clusters represent three distinct failure modes for safety-critical
infrastructure. \textbf{Opportunistic agents} (overload contribution 73.7\%)
drive chronic system overload: when nearly half of all agents request at rates
above 0.8, aggregate demand persistently exceeds capacity, making safe
operation impossible regardless of the behavior of the remaining agents.
\textbf{Conservative agents} experience starvation periods of up to 100
rounds, representing subsystems that go entirely offline---a silent failure
mode that does not trigger overload alerts but causes complete operational
downtime. \textbf{Aggressive agents}, despite having the best individual
efficiency (0.624), still contribute to overload at 20.6\% of rounds,
meaning they benefit personally from moderate requesting but impose
coordination costs on the system. The absence of Steady agents means there
is no self-regulating subpopulation that naturally damps demand variance;
every LLM agent, regardless of prompt, sorts into a failure mode that is
harmful at the system level.
\section{Discussion}

\subsection{Why LLMs Fail in Safety-First Resource Allocation}

The capacity-matching baseline is a strong benchmark for the \emph{capacity-matching} objective because it is calibrated to satisfy $\mathbb{E}[A]=C$ under the i.i.d.\ model and is intentionally unpredictable. However, it should not be interpreted as a universal optimum for every notion of ``safety'': if the sole objective is to minimize the overload probability $\mathbb{P}(A>C)$, one can always reduce overload by choosing $p<C/N$, at the expense of increased under-utilization (waste). Accordingly, we use $p=C/N$ as a calibrated reference point for safety-critical settings where both avoiding systematic over-demand and preserving unpredictability matter. Future work should compare against other adaptive baselines such as reinforcement learning agents, fictitious play, regret-matching algorithms, or crowd-anticrowd strategies~\cite{johnson1999} to more fully characterize the LLM performance landscape. LLM agents, by contrast, tend to infer patterns from short histories and share similar inductive biases, creating correlated demand spikes.

Three mechanisms are consistent with the observed outcomes. First, spurious pattern formation arises because short history windows (5--10 rounds) encourage overfitting to noise, leading to synchronized over-requesting when agents perceive a pattern (e.g., ``bar was empty last 2 rounds $\to$ probably empty now''). Second, shared priors and similar reasoning traces cause even AI agents with different ``personalities'' to converge to analogous heuristics such as trend-following and recent-average extrapolation. Third, decision determinism means that without explicit post-processing (e.g., $\varepsilon$-greedy), action distributions collapse, producing persistent unsafe regimes with low behavioral variance but high correlation.

\subsection{Implications for AI Infrastructure Safety}

The results suggest concrete guidance for practitioners deploying multi-AI-agent controllers in capacity-constrained loops. First, do not rely on pure LLM reasoning for real-time, safety-critical congestion control where overload is expensive or dangerous. Second, prefer calibrated stochastic policies (e.g., capacity-matching randomization) when unpredictability is safety-relevant. Third, use LLMs for high-level policy selection (e.g., dynamically setting $p_t$ or allocating priority classes) while preserving randomized low-level execution. Fourth, audit correlated failure modes: system-level risk emerges from AI agent similarity, not only from individual competence---``more intelligence'' can mean ``less safety.'' Finally, smaller models may be safer: for resource-constrained coordination, smaller models (53.8\% overload) reduce overload by 18.7 percentage points relative to larger models (72.5\%), i.e., about a 25.8\% relative reduction, suggesting task-model matching matters. This comparison should be validated with fully matched experimental conditions (Sec.~6.1).

Before deploying LLM agents in critical infrastructure, practitioners should consider whether the problem is resource-constrained with El Farol-type coordination, whether system overload is dangerous, and whether simple random strategies with calibrated $p=C/N$ would suffice. If the answer is yes, simple strategies may be preferable to sophisticated AI.

\subsection{Limitations and Future Work}

This study is intentionally small-scale to enable controlled analysis. Population sizes of $N\in\{3,4\}$ are far smaller than real infrastructures, and scaling to $N\gg 10$ (e.g., 100--1000 AI agents) is critical for practical relevance. The time horizon of 100 rounds may not capture long-run learning dynamics, and extensions to 1,000--10,000 rounds would test equilibration. Capacity is fixed in our setup, whereas real systems have time-varying capacity $C_t$ (e.g., renewable energy supply, dynamic network bandwidth); adaptive $C_t$ is an important extension.

AI Agents do not communicate in our experiments, and explicit coordination channels (AI agent-to-agent messages, auctions, queues) may help or worsen synchrony; hybrid architectures merit study. We tested a fixed prompt structure, and while extensive prompt optimization might improve outcomes, it risks overfitting to the specific setup. Finally, future work should evaluate hybrid controllers where LLMs set high-level parameters (e.g., $p_t$, priority allocations) while stochastic execution enforces safety.

\section{Conclusion}

In safety-critical resource allocation, the relevant objective is not merely individual payoff but maintaining system operation within capacity across changing operating points. Under this framing, a principled baseline is capacity-matching randomization $p(\GO)=C/N$, which ensures $\mathbb{E}[A]=C$ and yields predictable overload probabilities under binomial assumptions that depend on $(N,C)$ (e.g., 25.9\% for $(3,1)$ and 31.25\% for $(4,2)$; Appendix~A).

Across 43 experiments with 154 AI agent instances, LLM agents fail to improve safety or efficiency relative to this baseline and often perform substantially worse: large models achieve 72.5\% overload (2.32$\times$ the 31.25\% baseline for $N=4, C=2$), small models 53.8\% (1.72$\times$ baseline), though these comparisons should be interpreted cautiously given that model size was not the only variable differing across runs (see Sec.~6.1). Homogeneous populations at $N=3$, $C=1$ show highly divergent outcomes depending on personality type: All-Risk-Averse achieves 25.0\% overload (marginally below the capacity-matching target of 25.9\%, though the difference is not statistically significant; see Sec.~4.3), while All-Contrarian (70.0\%), All-Trend-Follower (81.0\%), and All-Optimist (89.0\%) substantially exceed even the coin-flip baseline (50.0\%, theoretical), spanning a 25.0\%--89.0\% range in overload rate.

Quantitative behavioral clustering reveals three emergent AI agent profiles—Opportunistic (48.1\%), Aggressive (27.3\%), and Conservative (24.7\%)---with large descriptive separability effect sizes ($\eta^2 \in [0.19, 0.74]$ for key behavioral metrics; see Sec.~4.5 for interpretive caveats). Critically, no Steady AI agents emerge among LLMs, contrasting sharply with the random baseline. These findings emphasize a practical AI safety lesson: for shared infrastructure under hard capacity constraints, stronger reasoning and richer internal models do not 
guarantee safer collective behavior.

When overload is dangerous in small-scale, capacity-constrained systems without communication channels, it may be safer to randomize rationally with $p=C/N$ than to reason deterministically about short-term patterns. For such settings, intelligence is powerful, but knowing when \emph{not} to use pattern-matching may be the most appropriate design choice.

\subsection{Methodological Considerations and Future Directions}

Several methodological refinements would strengthen future work in this domain:

\paragraph{Statistical rigor.} The model-size comparison should be extended to include confidence intervals, standard errors, and formal significance tests (e.g., two-sample t-tests) for the headline safety comparisons. While we report single-point overload percentages (72.5\%, 53.8\%) drawn from experimental logs, a more rigorous statistical framework with effect sizes and power analysis would strengthen claims about model-size effects across configurations.

\paragraph{Experimental matching and confound control.} The 43 runs vary $N$, $C$, model size, personality prompts, and exploration mechanism simultaneously. Future work should employ factorial designs with explicit matching: for instance, ensuring that large-model and small-model runs use identical $(N,C)$ pairs, personality distributions, and exploration mechanisms to isolate the effect of model capacity. Without such control, the model-size comparison may be confounded by unintended correlations between configuration choices and model assignments.

\paragraph{Baseline diversity and comparative benchmarks.} The capacity-matching random baseline ($p=C/N$) is optimal by construction for matching the \emph{mean} demand to capacity ($\mathbb{E}[A]=C$) under the i.i.d.\ binomial model, which may create a favorable comparison standard for that particular objective. Future work should include other adaptive baselines such as simple reinforcement learning agents (e.g., Q-learning with $\varepsilon$-greedy), fictitious play, regret-matching algorithms, or crowd-anticrowd strategies~\cite{johnson1999}. Comparing LLM performance against these established coordination mechanisms would provide a richer characterization of where LLMs succeed or fail relative to the broader landscape of multi-agent learning algorithms.

\paragraph{Safety vs. payoff objectives.} This paper focuses exclusively on capacity-matching ($p=C/N$) as the baseline, prioritizing system safety over individual payoff. However, the minority-game literature establishes that payoff-optimal strategies can diverge from safety-optimal ones, particularly when the objective is maximizing expected winners rather than minimizing overload frequency. Future work should explicitly test both baseline types and clarify which objective (system safety vs. individual welfare) is being optimized, following Jo et al.~\cite{jo2023} who demonstrate fundamental incompatibilities between allocation fairness and outcome fairness in resource allocation systems.

\paragraph{Prompt engineering and robustness.} The personality prompts used in this study are brief and informal (Table~\ref{tab:personalities}). Given known sensitivity of LLMs to prompt framing, rephrasing, and ordering, future work should systematically vary prompt structures to assess whether the observed behavioral clustering (Aggressive, Conservative, Opportunistic) represents robust emergent phenomena or artifacts of specific prompt wording. Prompt optimization may improve coordination outcomes, though care must be taken to avoid overfitting to the experimental setup.

\paragraph{Model versioning and reproducibility.} For full reproducibility, future work should report exact model version strings (e.g., specific GPT-4 checkpoint dates, Claude Sonnet version numbers) and temperature settings for all runs. The rapid pace of model updates means that behavioral results may not replicate with newer checkpoints, and version-specific documentation is essential for cumulative science in this domain.

\section*{Code and Data Availability}
Code available from the authors on reasonable request.


\appendix

\section{Random-Baseline Choice: $p(\GO)=C/N$}

This appendix formalizes the primary baseline used throughout the paper: \emph{capacity-matching randomization}, where each AI agent independently requests a unit with probability $p(\GO)=C/N$. The key point is that this baseline is used as a calibrated reference that matches expected demand to capacity ($\mathbb{E}[A]=C$) in a safety-oriented framing, not for individual payoff optimality.

\subsection{I.I.D. Bernoulli Model and Induced Binomial Attendance}

Let $X_i\in\{0,1\}$ be the action of AI agent $i$ in a given round, with
\begin{equation}
  X_i \sim \mathrm{Bernoulli}(p), \quad \text{independently for } i=1,\dots,N.
\end{equation}
The total demand (attendance) is
\begin{equation}
  A = \sum_{i=1}^N X_i.
\end{equation}
Then $A\sim\mathrm{Binomial}(N,p)$.

\subsection{Mean, Variance, and Standard Deviation}

Using linearity of expectation,
\begin{equation}
  \mathbb{E}[A] = \sum_{i=1}^N \mathbb{E}[X_i] = \sum_{i=1}^N p = Np.
\end{equation}
Since $\mathrm{Var}(X_i)=p(1-p)$ for Bernoulli variables and the $X_i$ are independent,
\begin{equation}
  \mathrm{Var}(A)=\sum_{i=1}^N \mathrm{Var}(X_i)=Np(1-p).
\end{equation}
The standard deviation is therefore
\begin{equation}
  \sigma_A = \sqrt{\mathrm{Var}(A)} = \sqrt{Np(1-p)}.
\end{equation}

\paragraph{Optional generalization: heterogeneous baselines.}
If AI agent $i$ uses $\mathrm{Pr}(\GO)=p_i$ independently (not necessarily identical), then $A=\sum_i X_i$ with $X_i\sim\mathrm{Bernoulli}(p_i)$ independent, and
\begin{equation}
  \mathbb{E}[A] = \sum_{i=1}^N p_i, \quad \mathrm{Var}(A) = \sum_{i=1}^N p_i(1-p_i).
\end{equation}
This reduces to the homogeneous case when $p_i\equiv p$ for all $i$.

\subsection{Capacity Matching: Choosing $p=C/N$}

The capacity-matching choice sets
\begin{equation}
  p = \frac{C}{N}.
\end{equation}
This yields
\begin{equation}
  \mathbb{E}[A] = N\cdot\frac{C}{N} = C,
\end{equation}
so the \emph{expected} aggregate demand matches the system capacity exactly. The variance becomes
\begin{equation}
  \mathrm{Var}(A)=N\cdot\frac{C}{N}\left(1-\frac{C}{N}\right)=C\left(1-\frac{C}{N}\right)=C\,\frac{N-C}{N},
\end{equation}
and the standard deviation is
\begin{equation}
  \sigma_A = \sqrt{C\left(1-\frac{C}{N}\right)}.
\end{equation}
This baseline is therefore directly aligned with safety-centric control goals: avoiding systematic over-demand ($\mathbb{E}[A]>C$) or under-demand ($\mathbb{E}[A]<C$).

\subsection{General Overload Probability}

The overload probability is defined as
\begin{equation}
  \mathbb{P}(\text{overload}) = \mathbb{P}(A>C) = 1-\sum_{a=0}^{C}\binom{N}{a}p^a(1-p)^{N-a}.
\end{equation}
For specific configurations, we compute this explicitly below.

\subsection{Explicit Overload Probability for $N=3, C=1$}

For $N=3$ and $C=1$, capacity matching gives $p=C/N=1/3$.
Attendance is $A\sim\mathrm{Binomial}(3,1/3)$. The overload event is $\{A>1\}$.

Compute
\begin{align}
  \mathbb{P}(A>1)
  &= 1-\mathbb{P}(A=0)-\mathbb{P}(A=1) \\
  &= 1-(2/3)^3 - \binom{3}{1}(1/3)(2/3)^2 \\
  &= 1-\frac{8}{27}-\frac{12}{27} \\
  &= \frac{7}{27} \approx 0.2593 \; (25.93\%).
\end{align}
This provides the safety benchmark for $N=3, C=1$ experiments.

\subsection{Explicit Overload Probability for $N=3, C=2$}

For $N=3$ and $C=2$, capacity matching gives $p=C/N=2/3$.
Attendance is $A\sim\mathrm{Binomial}(3,2/3)$. The overload event is $\{A>2\}$, which requires $A=3$.

Compute
\begin{align}
  \mathbb{P}(A>2)
  &= \mathbb{P}(A=3) \\
  &= \binom{3}{3}(2/3)^3(1/3)^0 \\
  &= (2/3)^3 \\
  &= \frac{8}{27} \approx 0.2963 \; (29.63\%).
\end{align}
This provides the safety benchmark for $N=3, C=2$ experiments, as referenced in Sec.~3.3.

\subsection{Explicit Overload Probability for $N=4, C=2$}

For $N=4$ and $C=2$, capacity matching gives $p=C/N=2/4=1/2$.
Attendance is $A\sim\mathrm{Binomial}(4,1/2)$. The overload event is $\{A>2\}$.

Compute
\begin{align}
  \mathbb{P}(A>2)
  &= \mathbb{P}(A=3)+\mathbb{P}(A=4) \\
  &= \binom{4}{3}(1/2)^3(1/2)^1 + \binom{4}{4}(1/2)^4 \\
  &= 4\cdot\frac{1}{16} + 1\cdot\frac{1}{16} \\
  &= \frac{5}{16} = 0.3125 \; (31.25\%).
\end{align}
This provides the safety benchmark for the primary model-size comparison ($N=4, C=2$) reported in Sec.~4.2.

We note the following coincidence with the coin-flip baseline.
For $(N,C)=(4,2)$, the capacity-matching probability $p=C/N=1/2$ coincides with an unbiased coin flip. This is a special property of configurations where $C=N/2$; for all other $(N,C)$ pairs tested in this paper (e.g., $(3,1)$ with $p=1/3$; $(3,2)$ with $p=2/3$), the capacity-matching baseline differs from a coin flip. The coincidence does not weaken the comparison: the baseline's theoretical justification is that $p=C/N$ sets $\mathbb{E}[A]=C$, which is a system-calibrated choice derived from the capacity constraint, whereas a coin flip ($p=0.5$) lacks any such grounding in general. That both happen to agree at $C=N/2$ simply means the capacity-matching principle yields the symmetric strategy in the symmetric configuration.

\subsection{Two Baselines, Two Objectives: Safety vs Payoff}

The correct choice of baseline depends on whether the objective is a \emph{capacity-matching} criterion (e.g., minimizing overcrowding frequency or minimizing $|A_t-C|$) or a \emph{minority-game payoff} criterion. Jo et al.~\cite{jo2023} formalize this distinction in contextual resource allocation, demonstrating that optimizing for individual welfare (payoff) can be fundamentally incompatible with system-level fairness and safety metrics.

In many formulations of minority games, the ``winning action'' is defined so that
\begin{equation}
  \text{GO wins if } A_t\le C,\qquad \text{STAY wins if } A_t>C,
\end{equation}
and AI agents receive positive payoff for matching the winning action. Importantly, maximizing expected payoff (or expected number of winners) is \emph{not} generally equivalent to minimizing $\mathbb{P}(A_t>C)$, and thus the payoff-optimal mixed strategy need not equal $C/N$.

The \textbf{capacity-matching baseline} uses $p = C/N$ with the objective of system safety and resource utilization. It is designed to match expected demand to capacity ($\mathbb{E}[A]=C$) and should be used when overload is dangerous or expensive, such as in energy grids, bandwidth limits, or cloud cluster saturation.

The \textbf{payoff-optimal baseline} sets $p\in\arg\max_p W(p)$, where $W(p)$ is the expected number of winners per round. Define the expected number of winners as
\begin{equation}
  W(p) = \sum_{a=0}^{C} \mathbb{P}(A=a) \cdot a + \sum_{a=C+1}^{N} \mathbb{P}(A=a) \cdot (N-a),
\end{equation}
because if $a\le C$, the $a$ agents who chose \GO\  win, and if $a>C$, the $N-a$ agents who chose \STAY\  win.

For $N=3$ and $C=1$, the expected number of winners per round becomes
\begin{align}
W(p)
&= \sum_{a=0}^{1} \mathbb{P}(A=a)\cdot a \;+\; \sum_{a=2}^{3} \mathbb{P}(A=a)\cdot (3-a) \nonumber\\
&= \mathbb{P}(A=0)\cdot 0 \;+\; \mathbb{P}(A=1)\cdot 1 \;+\; \mathbb{P}(A=2)\cdot 1 \;+\; \mathbb{P}(A=3)\cdot 0 \nonumber\\
&= \mathbb{P}(A=1) + \mathbb{P}(A=2) \nonumber\\
&= \binom{3}{1}p(1-p)^2 \;+\; \binom{3}{2}p^2(1-p) \nonumber\\
&= 3p(1-p)^2 + 3p^2(1-p) \nonumber\\
&= 3p(1-p) \nonumber\\
&= 3p - 3p^2.
\end{align}

This quadratic is maximized at $p=1/2$ (not $p=1/3$). Under this payoff-optimal choice with $p=0.5$, the overload probability is:
\begin{align}
  \mathbb{P}(A>1)
  &= 1-\mathbb{P}(A=0)-\mathbb{P}(A=1) \\
  &= 1-(1/2)^3-\binom{3}{1}(1/2)(1/2)^2 \\
  &= 1-\frac{1}{8}-\frac{3}{8} \\
  &= \frac{1}{2} = 50.0\%.
\end{align}
By contrast, under the safety-oriented capacity-matching choice $p=C/N=1/3$, we have $\mathbb{P}(A>1)=7/27\approx 25.9\%$. Thus payoff optimization increases overload risk relative to the safety-oriented baseline. In this paper, the core framing is safety-critical infrastructure, so the capacity-matching baseline is the correct reference point.

\subsection{Correlation Amplification in Shared-Reasoning Agents}

To formalize why correlated reasoning can increase overload risk, consider binary decisions
$X_i \in \{0,1\}$ with $\mathbb{E}[X_i] = p$ and $\mathrm{Var}(X_i) = p(1-p)$.
If agents are independent, aggregate demand
\[
A = \sum_{i=1}^{N} X_i
\]
has variance:
\[
\mathrm{Var}(A) = Np(1-p).
\]

However, if agents exhibit pairwise correlation
\[
\rho = \mathrm{Corr}(X_i, X_j), \quad i \neq j,
\]
then:
\[
\mathrm{Var}(A)
= \sum_{i} \mathrm{Var}(X_i)
+ \sum_{i \neq j} \mathrm{Cov}(X_i, X_j).
\]

Since
\[
\mathrm{Cov}(X_i, X_j) = \rho \, p(1-p),
\]
we obtain:
\[
\mathrm{Var}(A)
= Np(1-p) + N(N-1)\rho p(1-p),
\]
or equivalently,
\[
\mathrm{Var}(A)
= Np(1-p)\big[1 + (N-1)\rho\big].
\]

Thus, even small positive correlation $\rho > 0$ inflates aggregate variance multiplicatively by a factor $(1 + (N-1)\rho)$. 
Because positive pairwise correlation $\rho > 0$ inflates aggregate variance around the fixed mean $\mathbb{E}[A]=C$ by the multiplicative factor $(1+(N-1)\rho)$, a more dispersed distribution around the capacity threshold assigns higher probability mass to the overload tail $\{A>C\}$ under the capacity-matching mean condition, and correlated decision-making therefore directly amplifies congestion risk.

This provides a theoretical explanation for the empirical finding that LLM agents---sharing similar inductive biases and reasoning traces---exhibit higher overload rates than independent capacity-matching random agents. In congestion-control settings, unpredictability is not noise but a safety mechanism that suppresses correlation-driven variance inflation.

\makeatletter
\renewcommand{\@biblabel}[1]{[#1]}
\makeatother
\renewcommand{\refname}{References}

\end{document}